\documentclass[twocolumn,letterpaper]{IEEEAerospaceCLS}  

\usepackage[]{graphicx,float,latexsym,amssymb,amsfonts,amsmath,amstext,times}
\usepackage[colorlinks=true, urlcolor=blue, linkcolor=black, citecolor=black]{hyperref} 
\usepackage{amsmath}
\usepackage{amssymb}
\usepackage{pifont}
\usepackage{xcolor}
\usepackage{caption}
\usepackage{listings}
\usepackage{xcolor}
\usepackage{booktabs}
\usepackage{float}
\usepackage{arydshln}
\usepackage{pgfplots}
\pgfplotsset{compat=1.18}
\usepackage{tikz}
\usepackage{rotating}
\usepackage{dblfloatfix}

\lstdefinestyle{bashstyle}{
    language=bash,
    basicstyle=\ttfamily\footnotesize,
    breaklines=true,
    breakatwhitespace=true,
    columns=fullflexible,
    backgroundcolor=\color{gray!10},
    frame=single
}

\newcommand{\ignore}[1]{}  
\title{A Stochastic Approach to Terrain Maps\\for Safe Lunar Landing}

\author{%
Anja Sheppard\\
Draper Scholar\\
University of Michigan\\
2505 Hayward St.\\
Ann Arbor, MI 48109\\
anjashep@umich.edu
\and 
Chris Reale\\ 
The Charles Stark Draper Laboratory\\
555 Technology Square\\
Cambridge, MA 02139\\
creale@draper.com
\and
Katherine A. Skinner\\ 
University of Michigan\\
2505 Hayward St.\\
Ann Arbor, MI 48109\\
kskin@umich.edu
}

\begin{document}

\twocolumn[{
    \renewcommand\twocolumn[1][]{#1}
    \maketitle
    \centering
\includegraphics[width=0.98\linewidth]{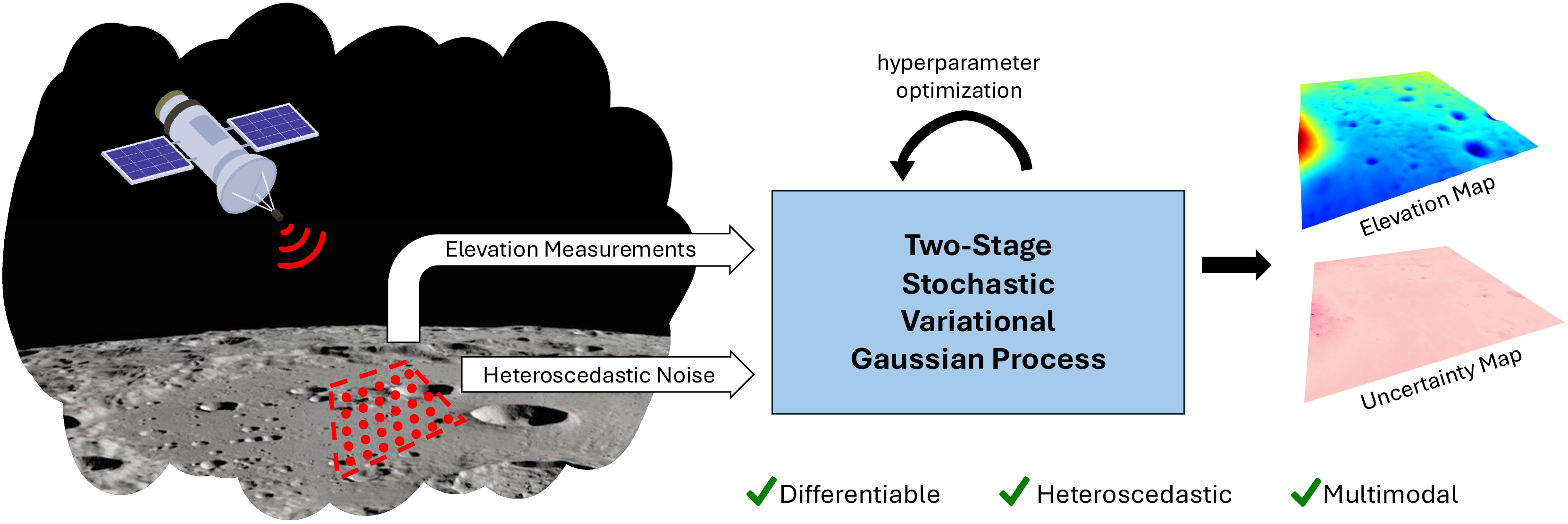}
    \captionof{figure}{An overview of our proposed two-stage Gaussian process (GP) framework for stochastic terrain maps.}
    \label{fig:overview}
    \vspace{1em}
}]

\begingroup
\renewcommand\thefootnote{}\footnotetext{\footnotesize 979-8-3315-7360-7/26/\$31.00~\copyright~2026 IEEE}%
\addtocounter{footnote}{-1}%
\endgroup

\begin{abstract}
Safely landing on the lunar surface is a challenging task, especially in the heavily-shadowed South Pole region where traditional vision-based hazard detection methods are not reliable. The potential existence of valuable resources at the lunar South Pole has made landing in that region a high priority for many space agencies and commercial companies. However, relying on a LiDAR for hazard detection during descent is risky, as this technology is fairly untested in the lunar environment. 

There exists a rich log of lunar surface data from the Lunar Reconnaissance Orbiter (LRO), which could be used to create informative prior maps of the surface before descent. In this work, we propose a method for generating stochastic elevation maps from LRO data using Gaussian processes (GPs), which are a powerful Bayesian framework for non-parametric modeling that produce accompanying uncertainty estimates. In high-risk environments such as autonomous spaceflight, interpretable estimates of terrain uncertainty are critical. However, no previous approaches to stochastic elevation mapping have taken LRO Digital Elevation Model (DEM) confidence maps into account, despite this data containing key information about the quality of the DEM in different areas.

To address this gap, we introduce a two-stage GP model in which a secondary GP learns spatially varying noise characteristics from DEM confidence data. This heteroscedastic information is then used to inform the noise parameters for the primary GP, which models the lunar terrain. Additionally, we use stochastic variational GPs to enable scalable training. By leveraging GPs, we are able to more accurately model the impact of heteroscedastic sensor noise on the resulting elevation map. As a result, our method produces more informative terrain uncertainty, which can be used for downstream tasks such as hazard detection and safe landing site selection. We compare against several stochastic mapping baselines using both simulated DEMs and real-world LRO Narrow Angle Camera data at the lunar South Pole.
\end{abstract} 

\tableofcontents

\section{Introduction}

Several recent failed attempts at landing on the lunar surface \cite{ispace2023hakuto}, \cite{astrobotic2024peregrine}, \cite{ispace2025hakuto} highlight the challenging nature of soft moon landings. Factors such as positioning error, radiation, sun angle, shadowing, surface albedo, and the lack of pre-existing high-resolution surface maps make this task particularly hard to accomplish.

The original Apollo moon landings relied on manual human piloting, which had a remarkable success rate \cite{lorenz2023planetary}. After a span of several decades, the Chinese Chang'e-3 lander successfully autonomously landed on the moon using a LiDAR for hazard detection and avoidance (HDA) \cite{yu2021technology}. Since then, interest from both commercial companies and national space agencies in achieving a moon landing has blossomed. Many of these recent approaches have used camera imagery and computer vision to detect hazards \cite{getchius2022hazard}, \cite{huang2022powered}, \cite{rallapalli2024landing}.

The lunar South Pole has recently been identified as a site of major interest due to a high likelihood of water presence under the surface \cite{flahaut2020regions}. Traditional computer vision approaches to hazard detection in this region are more challenging due to extreme shadowing. The alternative, real-time LiDAR scanning, is challenging to use during descent and has additional sources of uncertainty. Fuel is a limited resource aboard the landing spacecraft, and often only a few seconds are allocated for scanning, limiting the mappable area. In addition, spacecraft motion must be accurately estimated in order to create a global point cloud for analysis.

Existing 3D mapping frameworks for planning lunar landing sites typically include Digital Terrain Maps (DTMs) or Digital Elevation Maps (DEMs), which are georeferenced binned pointclouds at a fixed resolution \cite{jung2020}. However, these representations are discrete and non-differentiable, which pose a challenge for trajectory optimization. Although some stereo DEM matching tools output uncertainty \cite{beyer2018ames}, DEMs themselves do not support combining multiple sources of uncertainty or multiple sources of data at different resolutions.

Gaussian processes (GPs) have been proposed for terrain mapping, both in space \cite{tomita2023,hayner2025} and underwater \cite{torroba2022,song2023}. GPs excel at non-parametrically representing a continuous unknown function given noisy input data. However, they are computationally expensive in their standard formulation and they do not natively support spatially-dependent (heteroscedastic) noise \cite{williams2006gaussian}.

In this work we propose a heteroscedastic GP regression method for probabilistic lunar terrain mapping (see Fig. \ref{fig:overview}). Our approach -- a two stage stochastic variational GP framework -- explicitly allows for a more realistic noise model (spatially-dependent heteroscedastic noise) as well as scalability to large datasets. This enables more accurate terrain uncertainty outputs as compared to existing approaches. Additionally, we show quantitative and qualitative results on real-world Lunar Reconnaissance Orbiter (LRO) Narrow Angle Camera (NAC) stereo data as well as LuNaSynth simulated data.

\section{Related Works}

\subsection{Lunar Terrain Mapping Instruments}

The main source of orbital elevation data for lunar terrain is the LRO, which hosts the Lunar Orbiter Laser Altimeter (LOLA) and NAC sensors \cite{robinson2010}. LOLA emits pulses of five beams, which can provide a mapping resolution of five meters per pixel in certain areas. When an area of terrain has been imaged by several passes of LRO, higher resolution DEMs constructed from stereo pairs can be produced. The Kaguya (SELENE) mission hosted by JAXA has also produced stereo pairs for DEM construction, with resolutions ranging from 11 to 37 meters per pixel \cite{laura2024updated}.

The Ames Stereo Pipeline (ASP) is an open-source toolkit for taking orbiter stereo pairs and constructing DEMs \cite{beyer2018ames}. This toolkit can perform map projection, bundle adjustment, and shape from shading, while also propagating positional uncertainties and constructing heteroscedastic uncertainty maps from shadowing and albedo effects. 

Although maps constructed from LOLA data and ASP are useful in determining candidate landing sites, they are not of a high enough resolution to be completely relied upon during descent, as was the case for the Mars rover landings \cite{johnson2022}. However, the wealth of orbital data could still be of use for providing an \textit{a priori} terrain map with a corresponding uncertainty map to guide LiDAR imaging during descent. Due to a lack of existing frameworks for heteroscedastic probabilistic terrain maps, this use-case for orbital lunar elevation data has yet to be explored.

\subsection{Probabilistic Terrain Mapping}

Terrain mapping is a distinct task separate from 3D occupancy mapping, often referred to as 2.5D elevation mapping. There have been many approaches to representing 2.5D terrain information for the purpose of hazard detection, such as a standard gridded DEM \cite{liu2019planetary} or a DEM represented with Delauney triangulation \cite{jung2020}. DEMs do not support combining multipled data sources or different data source resolutions, which is where multi-resolution Laplacian pyramid decomposition \cite{schoppmann2021} and quadtrees \cite{setterfield2021lidar}, \cite{marcus2022landing} have contributed to the literature. However, quadtrees are discrete, deterministic, and not uncertainty-aware, which is challenging for optimization-based planning algorithms. 

As a more recent alternative, GPs provide a non-parametric, continuous representation of terrain with principled uncertainty estimates. One early method constructs a terrain map for a quadruped using a locally adaptive kernel \cite{plagemann2009bayesian}. Another approach builds a GP with extracted features from terrain images \cite{gao2018feature}, positing that areas with more detected features are more hazardous to a lander. For modeling terrain rather than hazards, one approach proposes a GP with an absolute exponential kernel \cite{tomita2024} while another approach proposes a GP with a radial basis function kernel \cite{hayner2025}. Another work constructs a GP from the LiDAR range measurements themselves in a spherical reference frame \cite{hansen2023range}.

There are several challenges with a standard GP approach--namely, a bottleneck in the number of training data points due to the need for a large covariance matrix and no model for heteroscedastic noise on the input or output training points. Several solutions have been proposed to alter GPs to overcome these challenges, such as the Stochastic Variational Gaussian Process (SVGP) \cite{hensman2015scalable}, which leverages inducing variables and batched stochastic optimization to make it possible to fit GPs to very large datasets. In a work that focuses on underwater probabilistic bathymetric mapping \cite{torroba2022}, the authors use an SVGP to represent noise on the vehicle pose as surveyed points are integrated into the map. Still, to the best of our knowledge, no current method has taken heteroscedastic noise on the elevation measurements into account. Our proposed method has a two-stage GP approach: one model represents the heteroscedastic noise, which is used to condition the likelihood of a second terrain model in order to produce a more realistic elevation uncertainty map.

There has been some prior work in defining how to incorporate heteroscedastic elevation noise into a GP. In \cite{goldberg1997}, the noise is assumed to be dependent on the inputs $\mathbf{X}$, and so another GP with Markov chain Monte Carlo for posterior sampling is used to model noise as a smooth function of the inputs. Another approach \cite{le2005} also assumes the noise is a random variable, and obtains a convex optimization problem through re-parameterization in order to properly solve for the noise. A more recent work aims to improve interpretability and uses polynomial regression-based noise modeling \cite{ozbayram2024}. All of these approaches seek to predict a noise function that can represent the heteroscedastic noise across different inputs. The simplest approach, as mentioned in \cite{lazaro2011}, is to model the heteroscedastic noise variance as another jointly-trained GP. However, this prior approach assumes no prior information is known about the heteroscedastic noise variance. As we have a variance prior, we modify this approach into a two-stage framework: first the noise GP is trained, and then it is frozen to be used as input to the terrain GP's likelihood.




\section{Preliminaries}

\subsection{Gaussian Processes}

Given a dataset of $n$ input points $\mathbf{X} = \{\mathbf{x}_i\}_{i=1}^n$ where  $\mathbf{x}_i \in \mathbb{R}^2$, and a set of $n$ corresponding samples $\mathbf{Y} = \{y_i\}_{i=1}^n$, we assume there is some underlying function $f(\mathbf{x})$, where $\mathbf{f}$ represents the true values of the function at $\mathbf{X}$. The sampled datapoints $\mathbf{Y}$ are assumed to have Gaussian white noise with standard deviation $\sigma$:
\begin{equation}
    y_i = f(\mathbf{x}_i) + \varepsilon, \;\;\;\;\; \varepsilon \sim \mathcal{N}(0, \sigma^2)
\end{equation}
This independent and identically distributed (i.i.d.) noise is assumed to be constant across the spatial dimension of the data, also known as homoscedasticity.

GPs are a powerful tool for non-parametric modeling \cite{williams2006gaussian}. A GP is a collection of random variables with a continuous time or space domain. It can also be thought of as a distribution over functions. A GP modeling the function $f$ is completely defined by a mean function $m(\mathbf{x})$ and a covariance function, or kernel $k(\mathbf{x}, \mathbf{x}')$:
\begin{equation}
    f \sim \mathcal{GP} \left(m(\mathbf{x}), k(\mathbf{x}, \mathbf{x}') \right)
\end{equation}
where
\begin{align}
    m(\mathbf{x}) &= \mathbb{E} [f(\mathbf{x})] \\
    k(\mathbf{x}, \mathbf{x}') &= \text{Cov}[f(\mathbf{x}), f(\mathbf{x}')].
\end{align}
The kernel encodes the covariance of datapoints in the GP. It is by definition positive semi-definite, symmetric, and invertible. Kernels can be stationary, meaning that the covariance is only dependent on the relative distance between $\mathbf{x}$ and $\mathbf{x}'$, or they can be non-stationary, meaning that the covariance is dependent on the value of the inputs themselves \cite{williams2006gaussian}. 

One common kernel, which will be discussed and used in this paper, is the radial basis function (also known as the squared exponential kernel):
\begin{equation}
    k_\text{RBF}(\mathbf{x}, \mathbf{x}') = \exp{\left(- \frac{\mid\mid \mathbf{x} - \mathbf{x}' \mid\mid^2}{2l^2} \right)}.
\end{equation}
Related to the radial basis function kernel is the rational quadratic kernel \cite{williams2006gaussian}, which is a scale mixture of many radial basis function kernels allowing for different lengthscales:
\begin{equation}
    k_\text{RQ}(\mathbf{x}, \mathbf{x}') = \left(1 + \frac{\mid\mid \mathbf{x} - \mathbf{x}' \mid\mid^2}{2 \alpha l^2} \right)^{-\alpha}.
\end{equation}
Additionally, there is the absolute exponential kernel, defined by \cite{tomita2023} as
\begin{equation}
    k_{AE}(\mathbf{x}, \mathbf{x}') = \exp{\left( -\frac{\mid\mid \mathbf{x} - \mathbf{x}' \mid\mid}{l} \right)}.
\end{equation}
The Mat\'ern kernel is defined as follows:
\begin{equation}
    k_M(\mathbf{x},\mathbf{x}')=
    \frac{2^{1-\nu}}{\Gamma(\nu)}
    \!\left(\frac{\sqrt{2\nu}\,\|\mathbf{x}-\mathbf{x}'\|}{l}\right)^{\!\nu}\!
    K_\nu\!\left(\frac{\sqrt{2\nu}\,\|\mathbf{x}-\mathbf{x}'\|}{l}\right)
\end{equation}
where $\nu$, controls the smoothness of the function, $\Gamma$ is the Gamma function, and $K_\nu$ is a modified Bessel function \cite{williams2006gaussian}.
Kernels can be summed and maintain their positive semi-definiteness. All three kernels employ the hyperparameter $l$, also known as the lengthscale, which defines the correlation distance between data points, effectively controlling how much influence neighboring data points have on each other. Typically, a scale kernel is added to these base kernels to allow for optimization of the output scale factor $\sigma_s$, which controls the magnitude of the predictions:
\begin{equation}
    k_{s}(\mathbf{x}, \mathbf{x}') = \sigma^2_s k_\text{base} (\mathbf{x}, \mathbf{x}')
\end{equation}
The set of hyperparameters for a specific kernel is designated by $\theta$. These are typically optimized via gradient descent by maximizing the log marginal likelihood \cite{gardner2018gpytorch}.

The Bayesian update process can be expressed using Bayes' Rule:
\begin{equation}
    p(\mathbf{f} \mid \mathbf{X}, \mathbf{Y}) = \frac{p(\mathbf{Y} \mid \mathbf{f}, \mathbf{X}) \; p(\mathbf{f})}{p(\mathbf{Y} \mid \mathbf{X})}.
\end{equation}
The prior, $p(\mathbf{f})$ is simply another GP, and the likelihood $p(\mathbf{Y} \mid \mathbf{f}, \mathbf{X})$ represents the model noise on the observed data. For a candidate point $\mathbf{x}^*$, the posterior predictive is defined as:
\begin{equation}
    p(f(\mathbf{x}^*) \mid \mathbf{X}, \mathbf{Y}) = \mathbf{N}(f(\mathbf{x}^*) \mid \mu_*, \sigma^2_*)
\end{equation}
where
\begin{align}
    \mu_* &= m(\mathbf{x}^*) + k(\mathbf{x}^*, \mathbf{X}) [k(\mathbf{X}, \mathbf{X}) + \sigma^2 I]^{-1} (\mathbf{Y} - m(\mathbf{X})) \\
    \sigma^2_* &= k(\mathbf{x}^*, \mathbf{x}^*) - k(\mathbf{x}^*, \mathbf{X}) [k(\mathbf{X}, \mathbf{X}) + \sigma^2 I]^{-1} k(\mathbf{X}, \mathbf{x}^*).
\end{align}
We overload notation so that $k(\mathbf{x}^*, \mathbf{X})$ denotes the vector $[k(\mathbf{x}^*, \mathbf{x}_1), \;\dots,\; k(\mathbf{x}^*, \mathbf{x}_n)]$ and $k(\mathbf{X}, \mathbf{X})$ denotes the Gram matrix with entires $k(\mathbf{x}_i, \mathbf{x}_j)$.

Using these equations, we can analytically compute the predictive posterior given inputs $\mathbf{X}$ and noisy random variable measurements $\mathbf{Y}$ where properties of the posterior process are determined by choice of kernel and hyperparameter optimization.

The standard GP assumes that the observed noise $\varepsilon$ on $\mathbf{Y}$ is homoscedastic, i.e. does not vary spatially across different data points. If we want to incorporate heteroscedastic sensor uncertainty to reflect spatially-varying DEM uncertainty from the terrain data, this formulation will not suffice. Similarly, a standard GP approach does not scale well to large batches of data.

\begin{figure*}[ht]
    \centering
    \includegraphics[width=\linewidth]{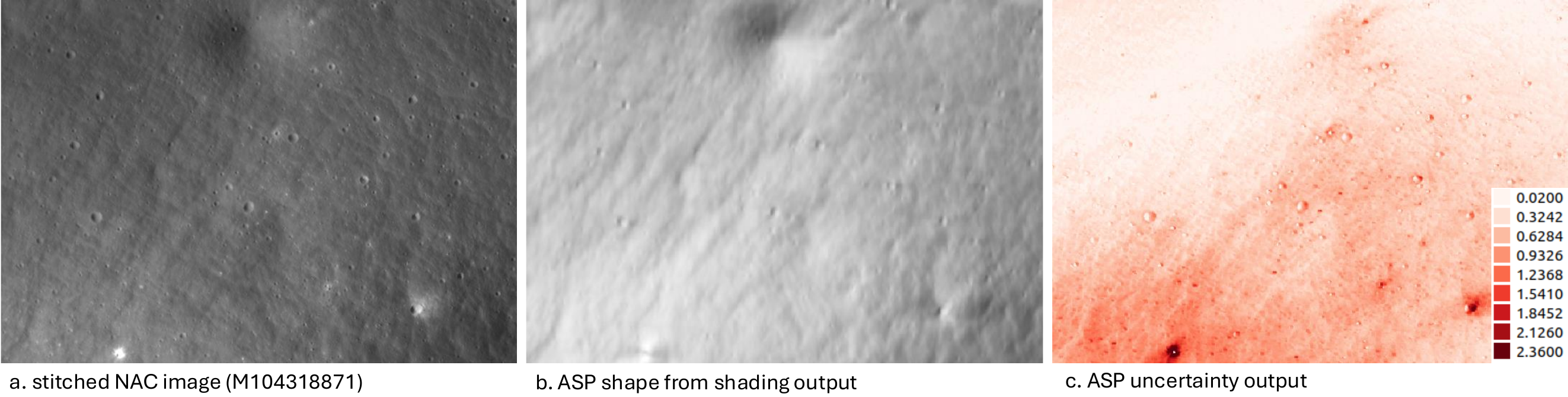}
    \caption{Data products from the ASP stereo matching process. From the left, a. is one of the images in the stereo pair (stitched left and right \texttt{M104318871}); b. is the hillshaded DEM output after map projection, stereo matching, and shape from shading; c. is the uncertainty map after shape from shading.}
    \label{fig:asp}
\end{figure*}

\subsection{Sparse and Stochastic Variational Gaussian Processes}

Large-scale batched training is impossible with a standard GP due to the computational bottleneck of the inversion of the $n \times n$ covariance matrix $k(\mathbf{X}, \mathbf{X})$ when computing the posterior. This requires $\mathcal{O}(n^3)$ time, which is prohibitive for large datasets. Sparse GPs \cite{titsias2009variational} mitigate this by introducing a set of $m \ll n$ inducing points, $\mathbf{Z} = \{\mathbf{z}_i\}_{i=1}^m$, associated with latent function values $\mathbf{u} = f(\mathbf{Z})$. 

The joint distribution over the original function values $\mathbf{f} = f(\mathbf{X})$ and the inducing variables $\mathbf{u} = f(\mathbf{Z})$ can be decomposed into the conditional and marginal priors:
\begin{equation}
    p(\mathbf{f}, \mathbf{u}) = p(\mathbf{f} \mid \mathbf{u}) \; p(\mathbf{u}).
\end{equation}

We introduce a joint variational distribution $q$:
\begin{equation}
    q(\mathbf{f}, \mathbf{u}) = p(\mathbf{f} \mid \mathbf{u}) \; q(\mathbf{u})
\end{equation}
Additionally, we introduce a second variational Gaussian distribution $q(\mathbf{u})$.
Then, we can obtain the marginal variational distribution over $\mathbf{f}$ by marginalizing out the inducing variables:
\begin{equation}
q(\mathbf{f}) = \int q(\mathbf{f} \mid \mathbf{u}) \, q(\mathbf{u}) \, d\mathbf{u} = \mathcal{N}(\mathbf{f} \mid \boldsymbol{\mu}_f, \mathbf{\Sigma}_f).
\end{equation}
Because both $p(\mathbf{f} \mid \mathbf{u})$ and $q(\mathbf{u})$ are Gaussian, this integral is analytically tractable, yielding a Gaussian distribution for $q(\mathbf{f})$. 

To approximate the true posterior $p(\mathbf{f} \mid \mathbf{Y})$ given $q(\mathbf{f})$ in a way that is tractable for large datasets, the evidence lower bound (ELBO) is used to optimize inducing points $\mathbf{Z}$ and kernel hyperparameters $\theta$ to make $q(\mathbf{f})$ as close as possible to the true posterior \cite{titsias2009variational}.

The first term of the ELBO objective (eq. \ref{eq:elbo}) is the expected log-likelihood of the observed data under the variational posterior. This is tractable if the likelihood is Gaussian. The second term $\text{KL}[q(\mathbf{u}) \mid\mid p(\mathbf{u})]$ is the Kullback-Leibler (KL) divergence between the variational distribution over inducing points and the GP prior over those inducing points, which acts as a regularizer.
\begin{equation}\label{eq:elbo}
\mathcal{L}_\text{ELBO}^\text{batch} = \frac{n}{b} \sum_{i\in\text{batch}} \mathbb{E}_{q(f_i)} \left[\log p(y_i \mid f_i) \right] - \text{KL}\left[q(\mathbf{u}) \mid\mid p(\mathbf{u}) \right].
\end{equation}
As introduced in \cite{hensman2013gaussian}, the expected log-likelihood can be approximated via stochastic optimization in mini-batches.

\begin{table*}[!h] 
    \centering 
    \renewcommand{\arraystretch}{1.2}
    \caption{Terrain mapping error and uncertainty calibrations on ASP NAC dataset and LuNaSynth synthetic data. For all three metrics, lower is better. Exact GP methods are compared only against other exact GPs, and similarly with variational GPs.} 
    \label{table:results} 
    \begin{tabular}{l|cccc|ccc} 
     \toprule 
        \specialrule{0.9pt}{0pt}{2pt} 
        && \multicolumn{3}{c|}{NAC Stereo} & \multicolumn{3}{c}{LuNaSynth \cite{restrepo2024technology}} \\
        Method && RMSE ($\downarrow$) & NLPD ($\downarrow$) & AUSE ($\downarrow$) &  RMSE ($\downarrow$) & NLPD ($\downarrow$) & AUSE ($\downarrow$) \\ 
        \specialrule{0.6pt}{1pt}{2pt} 
        \specialrule{0.6pt}{1pt}{2pt} 
        Tomita et al. \cite{tomita2023} && 0.0999 & -2.3007 & 0.0410 & 0.0240 & 3.4246 & 0.0081 \\
        Hayner et al. \cite{hayner2025} && 0.1079 & -2.3768 & 0.0451 & 0.0239 & 3.4245 & 0.0081\\ 
        Ours-Exact && \textbf{0.0976} & \textbf{-2.6418} & \textbf{0.0395} & \textbf{0.0009} & \textbf{-0.8367} & \textbf{0.0002} \\ 
        \addlinespace[1pt]\hdashline[3pt/2pt]\addlinespace[2pt]
        Torroba et al. \cite{torroba2022} && 0.3036 & -0.2592 & 0.1199 & 0.0242 & 3.4342 & 0.0089 \\ 
        Ours-Variational && \textbf{0.1472} & \textbf{-2.2166} & \textbf{0.0652} & \textbf{0.0011} & \textbf{0.2353} & \textbf{0.0004} \\ 
        \specialrule{0.9pt}{1pt}{2pt} 
    \end{tabular}
\end{table*}

\section{Technical Approach}

The highest resolution \textit{a priori} elevation data of the lunar surface comes from performing stereo estimation from NAC images. The ASP process propagates positional uncertainty from the camera poses and, after conducting shape from shading, outputs a heteroscedastic uncertainty map calculated by determining the height perturbation at each pixel that causes the simulated image to change by more than twice the difference between the unperturbed simulation and the measured image at that point \cite{beyer2018ames}. We display sample results from this process in Fig. \ref{fig:asp}, which shows a raw input NAC image; the output DEM after ASP's map projection, bundle adjustment, stereo matching, and shape from shading; and ASP's noise estimation which captures heteroscedastic noise for each input data point to the GP.

As we now have a prior on the heteroscedastic uncertainty, we formulate a two-stage framework with a separate noise GP and terrain GP. We modify our problem definition so that the output $y$ is:
\begin{equation}
    y_i = f(\mathbf{x}_i) + \varepsilon_i, \;\;\;\; \varepsilon_i \sim \mathcal{N}(0, \exp(g(\mathbf{x}_i)))
\end{equation}
and where $g(\mathbf{x}_i)$ is the log of the spatially-dependent noise variance (to ensure positivity):
\begin{equation}
    \log(r_i) = g(\mathbf{x}_i) + \zeta_i, \;\;\;\; \zeta_i \sim \mathcal{N}(0, \sigma_\zeta^2)
\end{equation}
We place a GP prior on both the unknown terrain function $f$ and the unknown noise covariance function $g$.
\begin{equation}
    f \sim \mathcal{GP} \left(m_f(\mathbf{x}), k_\text{RQ}(\mathbf{x}, \mathbf{x}') \right)
\end{equation}
\begin{equation}
    g \sim \mathcal{GP} \left(m_g(\mathbf{x}), k_\text{RBF}(\mathbf{x}, \mathbf{x}') \right)
\end{equation}
Additionally, we have a set of noisy samples $\mathbf{R} = \{r_i\}^n_{i=1}$ of the true noise variance function. Therefore, we can directly learn the process $g$ without jointly optimizing both GPs. We instead formulate a two-step approach where the noise GP is trained first and then frozen as an input to the terrain GP likelihood. The noise GP utilizes the radial basis function kernel to take advantage of its smoothing properties, while the terrain GP $f$ leverages the rational quadratic kernel for more variability in lengthscale. The terrain GP is initialized with a custom mean prior from the low resolution sample of the input data (see section \ref{sec:asp}).

We then express the likelihood of the observations $\mathbf{Y}$ given the terrain function values $\mathbf{f}$ and the noise function values $\mathbf{g} = g(\mathbf{X})$: 
\begin{equation}
    p(\mathbf{Y} \mid \mathbf{f}, \mathbf{g}) 
    = \prod_{i=1}^n \mathcal{N}\!\left(y_i \mid f(\mathbf{x}_i), \; \exp(g(\mathbf{x}_i)) \right),
\end{equation}
where we parameterize the variance process as $\exp(g(\mathbf{x}))$ to ensure positivity. With this likelihood, the posterior over $f$ is given by Bayes' rule: 
\begin{equation}
    p(\mathbf{f} \mid \mathbf{X}, \mathbf{Y}, \mathbf{g}) 
    \propto p(\mathbf{Y} \mid \mathbf{f}, \mathbf{g}) \, p(\mathbf{f} \mid \mathbf{X}),
\end{equation}
where the prior $p(\mathbf{f} \mid \mathbf{X})$ is the GP prior defined by $(m_f, k_\text{RQ})$. Since we condition on the posterior mean from the noise GP -- which gives us fixed heteroscedastic noise variances and a Gaussian likelihood -- the posterior over the terrain function remains analytically tractable.

However, as previously mentioned, exact analytical GPs are very computationally intensive. We therefore also formulate a stochastic variational GP with known heteroscedastic noise:  
\begin{equation}
    q(\mathbf{f}) \approx p(\mathbf{f} \mid \mathbf{X}, \mathbf{Y}, \mathbf{g}).
\end{equation}
For the variational approach, we modify the expected log-likelihood term in the ELBO (see eq. \ref{eq:elbo}) to incorporate the learned heteroscedastic noise:
\begin{equation}
\mathbb{E}_{q(f_i)}[\log p(y_i \mid f_i, g_i)] = \mathbb{E}_{q(f_i)}[\log \mathcal{N}(y_i \mid f_i, \exp(g_i))].
\end{equation}
At prediction time, for a new test point $\mathbf{x}^*$, we first compute the noise GP posterior mean $\mu_g(\mathbf{x}^*)$, then use this as the log-variance in the terrain GP predictions. This two-stage approach reduces computational complexity from $\mathcal{O}(n^3)$ for exact GPs to $\mathcal{O}(m^3)$ for variational GPs with $m \ll n$ inducing points, enabling scalable terrain mapping with spatially-varying uncertainty quantification.

The GPs are constructed and trained in GPyTorch \cite{gardner2018gpytorch}, a popular Torch-based Python library for GPs. We z-score normalize both the inputs $\mathbf{X}$ and outputs $\mathbf{Y}, \mathbf{R}$ to avoid numerical instabilities.
The inducing points are initialized from a random subset of the training inputs $\mathbf{X}$.

\section{Experiments and Discussion}

\begin{figure*}[!h]
    \centering
    \includegraphics[width=0.84\linewidth]{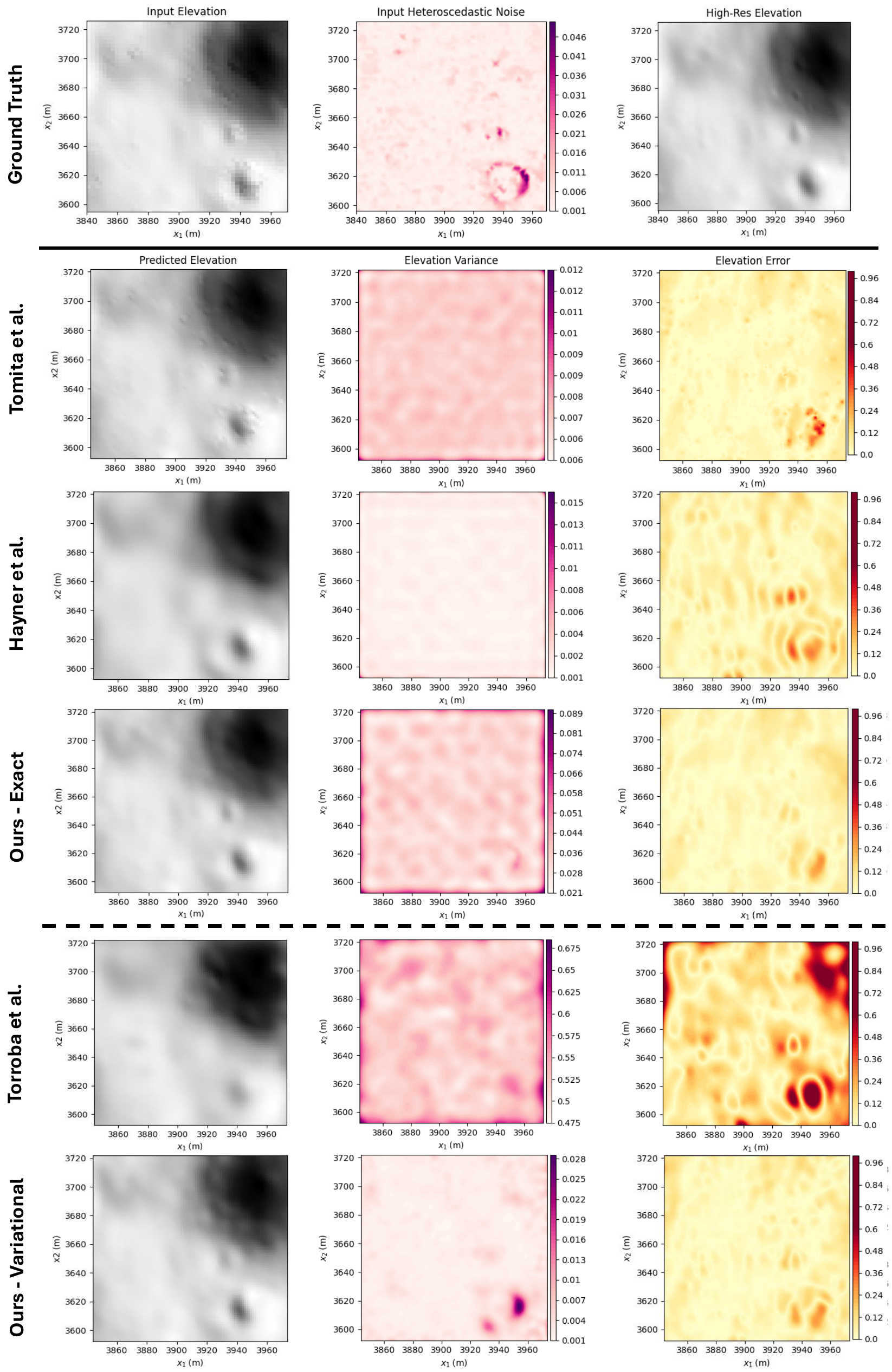}
    \caption{Qualitative comparison of our method against the three baselines on a tile from the dataset. Notice that the areas of higher uncertainty (middle column) match the areas of higher error (rightmost column) for our method.}
    \label{fig:qual}
\end{figure*}

\subsection{Ames Stereo Data Pipeline}\label{sec:asp}

We leverage the ASP for processing stereo pairs into DEMs and corresponding uncertainty maps \cite{beyer2018ames}. Exact commands for image processing are included in Appendix \ref{app:asp}.

Although NAC captures two images per sample, these do not have a large enough baseline for use as a stereo pair, hence the need for a second set of images taken of the same region but on a different orbit. We first stitch together the left and right NAC image pair for both sets of images in the stereo pair. Then we perform map projection, which projects the stereo pair onto a pre-existing low-resolution terrain map. This aids in reducing computation time during stereo matching and also produces better results in regions with steep drop offs or shadowing. To get the low-resolution terrain map for this step, we run stereo matching on the left and right images at a 40x lower resolution. Finally, to produce a higher quality map from albedo and shadowing in the input images, we run shape from shading. This also produces a heteroscedastic height errors map.

The resolution of this final map is about 1.3 meters per pixel. There is no source of ground truth elevation data for the moon at a higher resolution, so we consider this high resolution result from ASP our reference terrain. For training the terrain maps, we downsample the DEMs by a factor of two. To produce an even lower resolution prior, we downsample the high resolution maps by a factor of five. Another data source, such as LOLA elevation data, could also be used as the prior mean function. Additionally, we inject white noise onto the downsampled DEM according to the noise variance in the uncertainty plot. This allows us to simulate having a ``noise-free" reference trajectory in the high-resolution DEM.

\begin{figure*}[h]
    \centering
    \includegraphics[width=\textwidth]{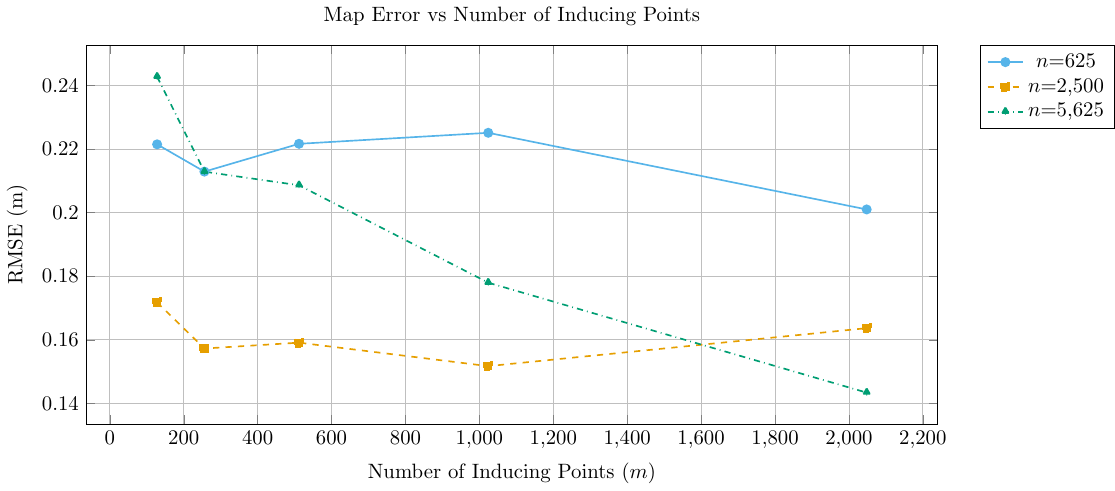}
    \caption{A comparison of the variational GP efficiency/accuracy tradeoff for our method. For each size dataset ($n$), there is an optimal number of inducing points that minimizes error while reducing computational load.}
    \label{fig:myplot}
\end{figure*}

\subsection{LuNaSynth Dataset}

LuNaSynth (previously known as LuNaMaps) is a synthetic data generation pipeline developed by NASA Ames for lunar navigation development \cite{restrepo2024technology}. The pipeline takes in a DEM as input, and procedurally generates rocks, regolith, and craters, as well as rendered images of the terrain under chosen illumination parameters. We use LuNaSynth to generate synthetic data that is of a higher resolution than DEMs produced by ASP. We downsample the high resolution DEMs by a factor of 2 for training the terrain maps. LuNaSynth does not model sensor noise, so we generate a sensor noise map from detected shadowed regions in the rendered images, and then randomly perturb the elevation measurements in our training data at pixels with higher uncertainty.

\subsection{Baselines}
We compare our method against three baselines: a standard GP with a radial basis function kernel \cite{hayner2023}, a standard GP with an absolute exponential kernel \cite{tomita2023}, and an SVGP with a Mat\'ern kernel \cite{torroba2022}. The standard GPs train quickly on small datasets and capture model uncertainty, but do not scale well to large datasets. The SVGP approach uses a variational framework to approximate the GP posterior, but it is not able to take heteroscedastic sensor noise into account. We train our method in two variants: an exact two-stage GP and variational two-stage GP. This allows us to compare against both sets of baselines and also see the strengths and limitations of the standard versus variational GP formulations.

There is open-source code for the SVGP bathymetric mapping approach, which we adapt for our data inputs \cite{torroba2022}. Although there is no code available for the other two methods \cite{tomita2023,hayner2023}, we use a standard GPyTorch \cite{gardner2018gpytorch} implementation according to the parameters described in the papers.

\subsection{Metrics}
We use three metrics to compare the GP-based terrain mapping methods. For evaluation, we sample the GP at 2x the resolution of the training data, matching the high resolution held-out test DEM. The first metric is the commonly-used Root Mean Square Error (RMSE), which evaluates how metrically accurate the resulting map is. This metric does not take the uncertainty into account.

The Negative Log Probability Density (NLPD) measures the error between a model's predictions and the ground truth, while also taking uncertainty into account \cite{williams2006gaussian}. It is the negative log likelihood of the test data ($y_i$) given the predictive distribution. A low NLPD indicates that the model is both accurately predicting the terrain and the uncertainties are calibrated.
\begin{equation}
    \text{NLPD} = - \frac{1}{n} \sum_{i=1}^n \log{p(y_i = \hat{y}_i \mid \mathbf{x}_{*_i}, \mu_{*_i}, \sigma^2_{*_i})}
\end{equation}

The Area Under the Sparsification Error curve (AUSE) \cite{ilg2018uncertainty} measures the quality of a model’s predictive uncertainty calibration by comparing the error reduction achieved when removing predictions with the highest estimated uncertainty against the error reduction when removing predictions based on the true error. We first define the model sparsification curve as the Mean Absolute Error (MAE) when a fraction $\alpha$ of the most uncertain predictions are removed from evaluation. Serving as the optimal baseline, the oracle sparsification curve is the MAE when a fraction $\alpha$ of the predictions with the highest error are removed from evaluation. The sparsification error is then defined as the difference between the two curves. The AUSE is the area under this sparsification error curve, integrated across all sparsification fractions $\alpha \in [0,1]$:
\begin{equation}
\text{AUSE} = \int_0^1 \text{Err}_\text{uncertainty}(\alpha) - \text{Err}_\text{oracle}(\alpha) \; d\alpha.
\end{equation}
A lower AUSE indicates that the predicted uncertainties better align with the true errors, meaning the model is well-calibrated for identifying which predictions are more reliable. We compute AUSE with discretizations of 50 datapoints.

\begin{table*}[!b] 
    \centering 
    \renewcommand{\arraystretch}{1.1}
    \caption{Training parameters for all five GP methods.} 
    \label{table:param} 
    \begin{tabular}{lccccccc} 
     \toprule 
        \specialrule{0.9pt}{0pt}{2pt} 
          && Tomita & Hayner & Ours-Exact && Torroba & Ours-Variational \\
        \specialrule{0.6pt}{1pt}{2pt}
        Learning Rate && 0.1 & 0.1 & 0.1 && 0.1 & 0.05 \\
        Epochs && 40 & 50 & 30 && 75 & 40 \\
        Batch Size && - & - & - && 256 & 256 \\
        Num. Inducing && - & - & - && 1024 & 1024 \\
        Kernel && AbsExp & RBF & RQ && Mat\'ern & RQ \\
        \specialrule{0.9pt}{1pt}{2pt} 
    \end{tabular}
\end{table*}

\subsection{Accuracy and Calibration Experiment}
We train each method on 961 ASP terrain tiles and 500 LuNaSynth terrain tiles. The hyperparameters are tuned during the optimization process, and all methods are supplied with the same hyperparameter priors. Loss convergence was used as the indicator for the number of epochs to train each method. Training took place on an Intel Xeon W-2245 8-core CPU, running Ubuntu 20.04, equipped with an NVIDIA GeForce RTX 3090 GPU (24GB VRAM). More training parameters are detailed in Appendix \ref{app:param}.

Quantitative results are detailed in Table \ref{table:results}. The methods are grouped by whether they are an SVGP or exact GP approach, as these are not directly comparable due to the variational GPs using approximations to reduce computational load. The exact GP variant of our approach outperforms both Tomita \cite{tomita2023} and Hayner \cite{hayner2025} baselines in terms of RMSE, NLPD, and AUSE. This indicates both accurate maps and well-calibrated variances. Additionally, the variational version of our approach outperforms the SVGP baseline \cite{torroba2022} in all three metrics on the real-world dataset and the synthetic dataset.

In Fig. \ref{fig:qual} we show a sample tile from the ASP dataset with the terrain map (leftmost column) and uncertainty map (center column) for each method. We see that Tomita's \cite{tomita2023} absolute exponential kernel captures noisy and sharp terrain edges, while Hayner's \cite{hayner2025} RBF kernel is more smooth. Torroba's SVGP approach \cite{torroba2022} has both higher error and a less expressive uncertainty map than our proposed approach, potentially due to the selection of the Mat\'ern kernel. Both variations of our method carry over the prior heteroscedastic noise information to the final variance plot, which more accurately reflects where the terrain process has higher error.

\subsection{Inducing Points Experiment}

In this experiment (see Fig. \ref{fig:myplot}), we vary the number of inducing points in the variational version of our approach and compare performance over multiple variants of the dataset with different sizes $n$. Through this experiment, we see that the number of inducing points has a strong impact on the accuracy of the final map, and more inducing points is not always necessarily better. There is a critical three-way trade-off between the number of inducing points, computational accuracy, and posterior prediction error. Selecting a variational approach allows a GP to scale to larger datasets.

\section{Conclusion}

In this work, we present a heteroscedastic GP for lunar terrain mapping that incorporates prior information about spatially-varying noise. Our two-stage training paradigm allows for the incorporation of prior uncertainty on the input terrain points, whereas existing approaches cannot leverage this valuable spatially-varying uncertainty data. We show quantitative metrics on two datasets -- one real and one simulated -- and on three baseline methods \cite{tomita2023,hayner2023,torroba2022} in comparison to our approach. Across RMSE, NLPD, and AUSE, our method displayed high terrain accuracy and well-calibrated uncertainty due to the incorporation of the heteroscedastic noise prior.

In the future, we plan to further develop our proposed two-stage variational heteroscedastic GP by demonstrating its use-case in downstream tasks such as hazard detection and avoidance. Specifically, we hope to include the terrain uncertainty in a hazard detection strategy. An additional future direction is the exploration of lander trajectory optimization given heteroscedastic terrain uncertainty provided by our approach.

\acknowledgements
This work was supported by NSF Grant No. DGE 2241144 and a Draper Scholars Fellowship. The authors would like to thank Ted Steiner for his valuable insights that helped shape the early stages of this work.

\appendices{}              

\section{Ames Stereo Pipeline Processing}\label{app:asp}

Below we present the specific commands and parameters utilized to generate our real-world dataset from the Ames Stereo Pipeline. All steps were conducted on an Intel Xeon W-2245 8-core CPU, running Ubuntu 20.04, equipped with an NVIDIA GeForce RTX 3090 GPU (24GB VRAM). The four DEMs can be downloaded from NASA PDS according to their identifiers as shown below.

\subsection{Stitching and Mapprojection}
This step runs \texttt{lronac2isis}, \texttt{lronaccal}, \texttt{lronacecho}, \texttt{spiceinit}, \texttt{noproj}, and \texttt{handmos} to create a stitched unprojected image for each pair.
\begin{lstlisting}[style=bashstyle]
lronac2mosaic.py M104318871LE.img M104318871RE.img
lronac2mosaic.py M104311715LE.img M104311715RE.img
\end{lstlisting}

Then, we use the intermediate mapproject step in order to improve the changes of pixel matching in areas of steep/complex terrain. Projecting the left and right stereo images onto an existing, lower-resolution terrain model may make the stereo matching step more likely to succeed.

\begin{lstlisting}[style=bashstyle]
parallel_stereo left.cub right.cub            \
    --subpixel-mode 1                           \
    run_coarse/run

point2dem --stereographic --auto-proj-center \
    --tr 40.0 --search-radius-factor 5 \
    run_coarse/run-PC.tif

dem_mosaic                  \
    --fill-search-radius 25 \
    --fill-power 8          \
    --fill-percent 10       \
    --fill-num-passes 3     \
    run_coarse/run-dem-DEM.tif -o run_coarse/run-smooth.tif

mapproject run_coarse/run-smooth.tif \
    M104318871LE.cub left_proj.tif

mapproject --ref-map left_proj.tif  \
    run_coarse/run-smooth.tif \
    M104311715LE.cub right_proj.tif
\end{lstlisting}

\subsection{Stereo Matching}
Next, we run the core stereo matching algorithm.

\begin{lstlisting}[style=bashstyle]
parallel_stereo                \
    --stereo-algorithm asp_mgm   \
    --subpixel-mode 9            \
    --sgm-collar-size 256        \
    left_proj.tif right_proj.tif \
    M104318871LE.cub M104311715LE.cub           \
    run_map/run                  \
    run_nomap/run-smooth.tif

point2dem -r moon --stereographic --proj-lon 0 --proj-lat -90 run_map_ba/run-PC.tif
\end{lstlisting}

There is an optional (but highly recommended) step here to use \texttt{gdal\_translate} to crop the DEM to a smaller region of interest before performing shape from shading. Shape from shading improves the final DEM quality by utilizing the albedo and shadowing on the images themselves.

\begin{lstlisting}[style=bashstyle]
sfs -i run_map_ba/run_crop_crop_PC.tif \   
    M104318871LE.cub M104311715LE.cub \
    --use-approx-camera-models \
    --crop-input-images \
    --reflectance-type 1 \
    --smoothness-weight 0.08  \
    --initial-dem-constraint-weight 0.001 \
    --max-iterations 10 \
    -o run_map_ba_sfs/run

parallel_sfs --estimate-height-errors \
    -i run_map_ba_bigsfs/run-DEM-final.tif \ 
    -o bigsfs_error/run \
    M104318871LE.cub M104311715LE.cub
\end{lstlisting}

\section{Training Parameters}\label{app:param}
In Table \ref{table:param}, we present the relevant GP training parameters for each of the methods evaluated in this paper. These parameters were the same across both datasets.


\bibliographystyle{IEEEtran}
\bibliography{references}

\thebiography

\begin{biographywithpic}
{Anja Sheppard}{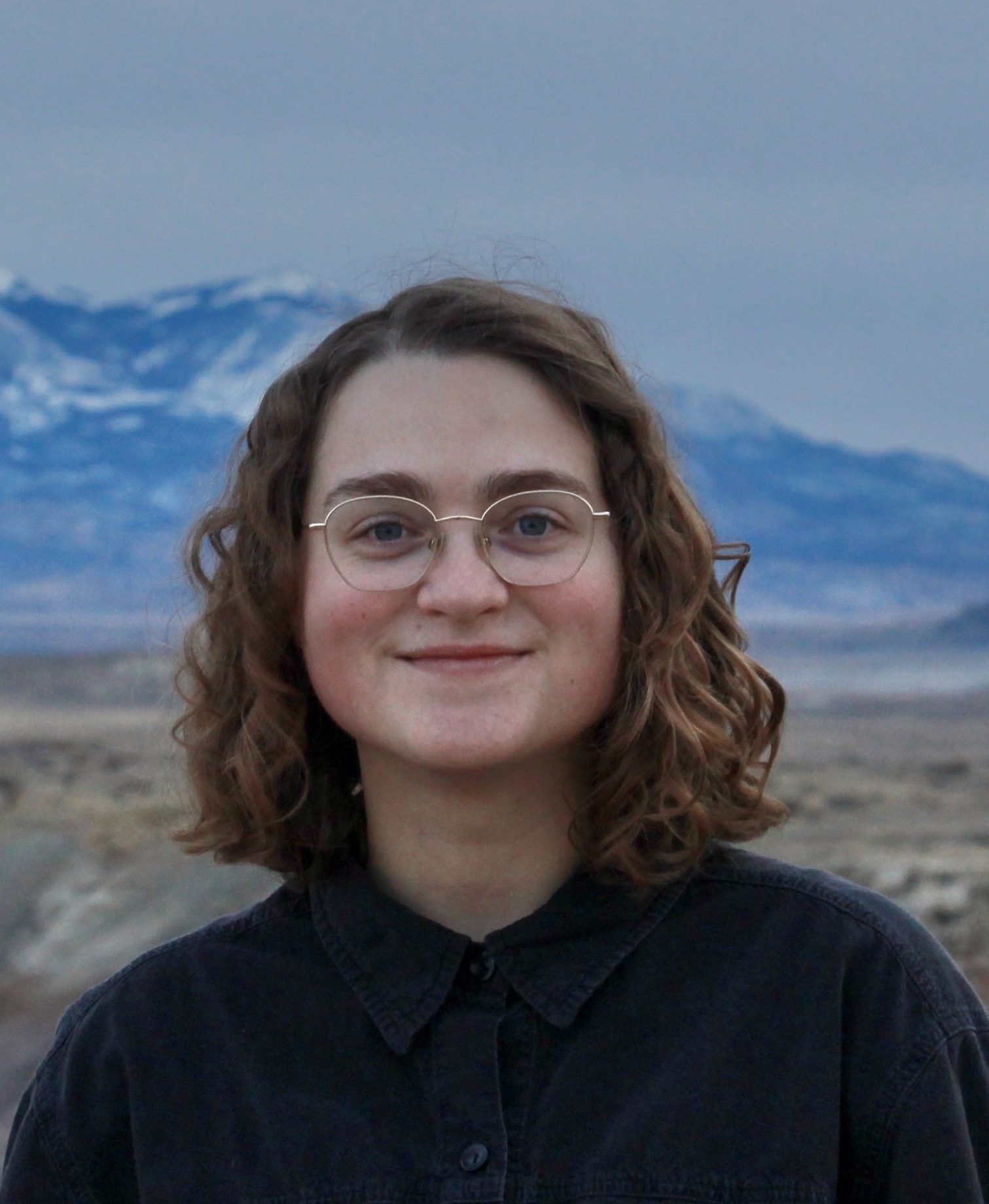} is a PhD Candidate in Robotics at the University of Michigan Field Robotics Group. She received an M.S. in Robotics at the University of Michigan and a B.S. in Computer Science from the University of Texas at Dallas. Her research focuses on probabilistic methods for multi-modal robot perception, both in space and underwater. Prior to her PhD, Anja contributed to NASA projects such as the Valkyrie humanoid robot and the ISS/Gateway space stations. She is a currently an NSF GRFP Fellow and a Draper Scholar in the Draper Embedded Perception and Machine Learning Group.
\end{biographywithpic}
\begin{biographywithpic}
{Chris Reale}{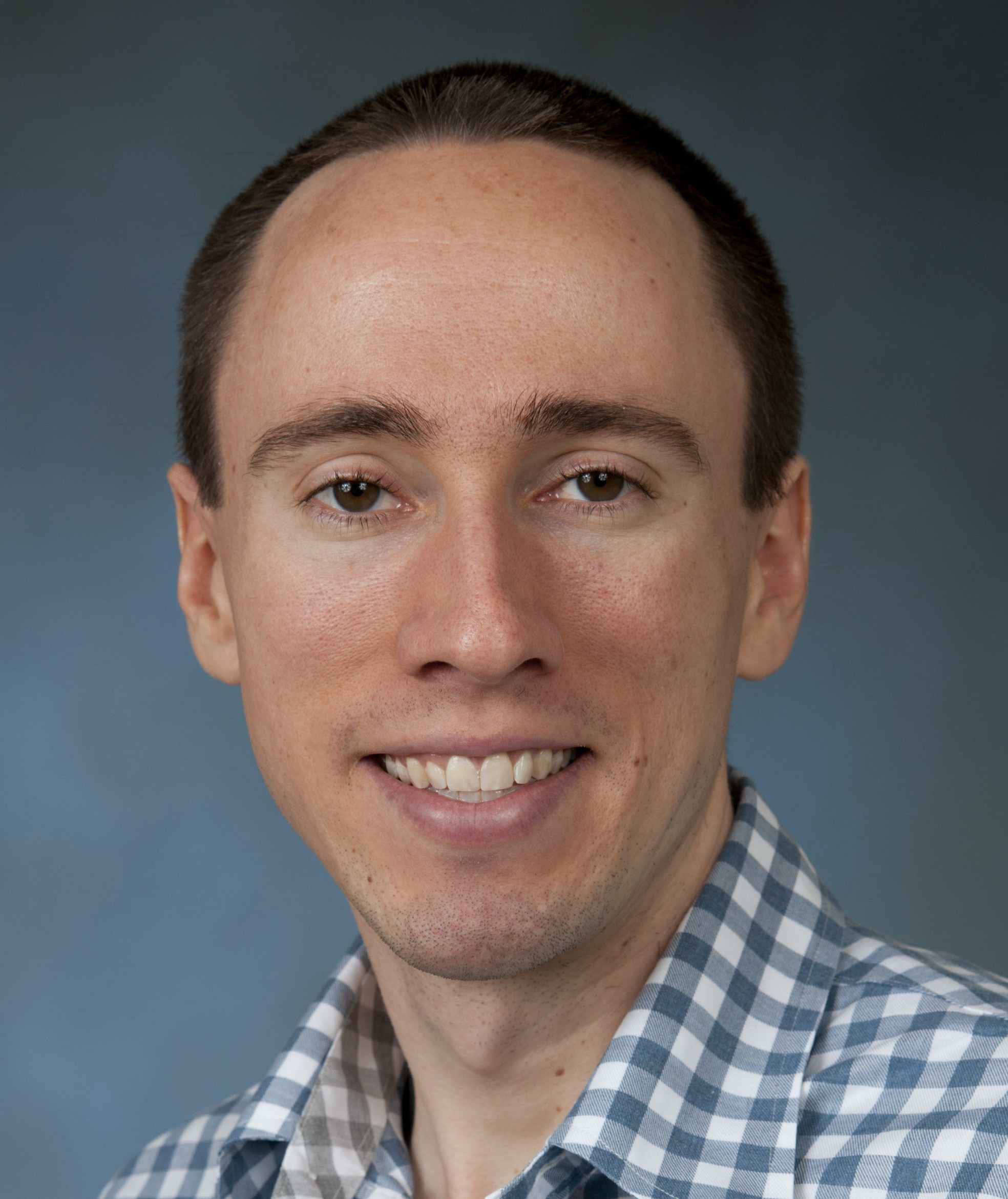} is a Principal Member of the Technical Staff in the Perception and Embedded Machine Learning group at Draper. He received his Ph.D. and M.S. from the University of Maryland and B.S. from Washington University in St. Louis, all in Electrical Engineering. He specializes in computer vision and machine learning. At Draper he has worked on applying computer vision and machine learning algorithms to solve problems in perception, navigation, autonomy, and calibration.
\end{biographywithpic}
\begin{biographywithpic}
{Katherine A. Skinner}{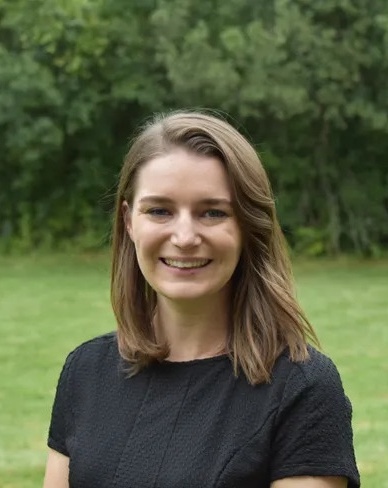} is an Assistant Professor in the Department of Robotics at the University of Michigan. She holds a courtesy appointment in the Department of Naval Architecture and Marine Engineering. Prior to this appointment, she was a Postdoctoral Fellow in the Daniel Guggenheim School of Aerospace Engineering and the School of Earth and Atmospheric Sciences at Georgia Institute of Technology. She received an M.S. and Ph.D. from the Robotics Institute at the University of Michigan, and a B.S.E. in Mechanical and Aerospace Engineering with a Certificate in Applications of Computing from Princeton University. She is a recipient of the NSF CAREER Award, ONR Young Investigator Award, and IEEE Robotics and Automation Letters Best Paper Award.

\end{biographywithpic}

\end{document}